\pdfoutput=1

\documentclass[11pt]{article}

\usepackage[final]{acl}

\usepackage{times}
\usepackage{latexsym}

\usepackage[T1]{fontenc}

\usepackage[utf8]{inputenc}

\usepackage{microtype}

\usepackage{inconsolata}

\usepackage{graphicx}

\usepackage{xspace}
\usepackage{booktabs}
\usepackage{adjustbox}
\usepackage{comment}
\usepackage{amsmath}
\usepackage{array}
\newcommand{\model}[0]{\textsc{SentenceSmith}\xspace}

\newcommand{\edge}[3]{\texttt{<#1,#2,#3>}}
\newcommand{\embedder}[1]{\texttt{#1}}

\title{Sentence Smith: Controllable Edits for Evaluating Text Embeddings}
 \author{Hongji Li ~~~ Andrianos Michail ~~~ Reto Gubelmann ~~~ Simon Clematide ~~~ Juri Opitz
 \\
         University of Zurich \\
           \small{
    \textbf{Correspondence:} \texttt{opitz.sci@gmail.com}
  }}

\begin{document}
\maketitle
\begin{abstract}
Controllable and transparent text generation has been a long-standing goal in NLP. Almost as long-standing is a general idea for addressing this challenge: Parsing text to a symbolic representation, and generating from it. However, earlier approaches were hindered by parsing and generation insufficiencies. Using modern parsers and a safety supervision mechanism, we show how close current methods come to this goal. Concretely, we propose the \model framework for English, which has three steps: 1.\ Parsing a sentence into a semantic graph. 2.\ Applying human-designed semantic manipulation rules. 3.\ Generating text from the manipulated graph. A final entailment check (4.) verifies the validity of the applied transformation. To demonstrate our framework's utility, we use it to induce hard negative text pairs that challenge text embedding models. Since the controllable generation makes it possible to clearly isolate different types of semantic shifts, we can evaluate text embedding models in a fine-grained way, also addressing an issue in current benchmarking where linguistic phenomena remain opaque. Human validation confirms that our transparent generation process produces texts of good quality. Notably, our way of generation is very resource-efficient, since it relies only on smaller neural networks. 
\end{abstract}

\section{Introduction}

\begin{figure}[ht]
\centering
\includegraphics[width=1.0\linewidth]{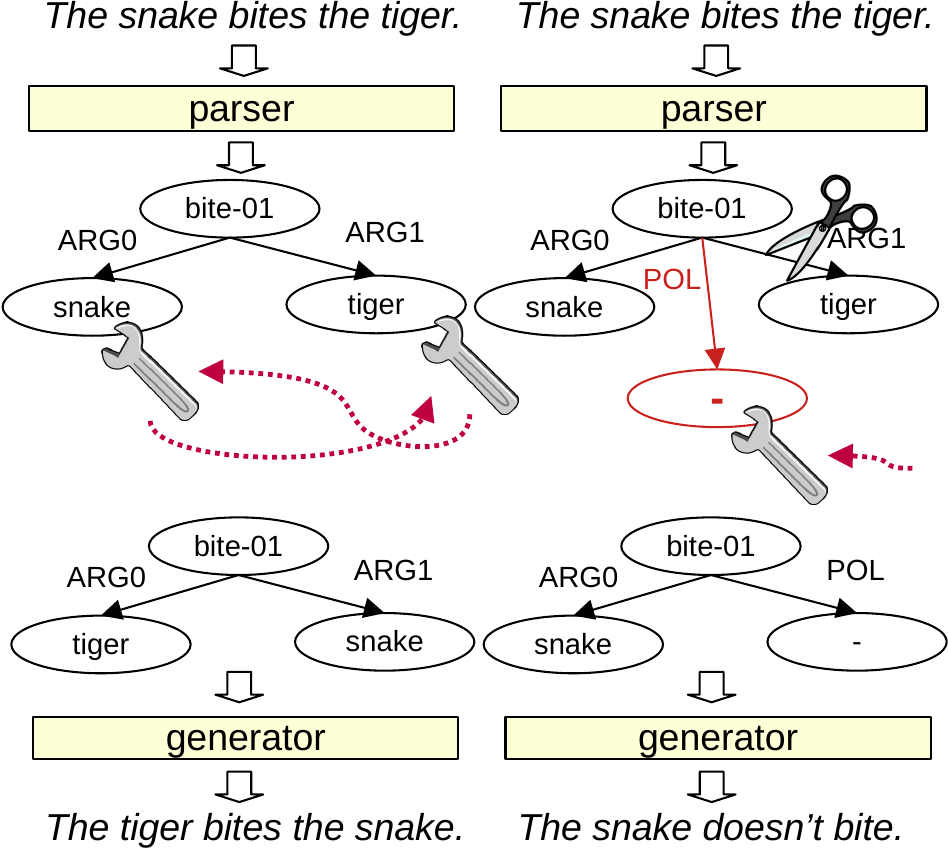}
\caption{Example application of two controlled text transformations. \textbf{Left}: Switching semantic roles in a sentence, where the agent (ARG0: the biter) becomes the patient (ARG1: the bitten), and vice versa. \textbf{Right}: Deleting the patient and negating the main predicate.}
\label{fig:srl-switch}
\end{figure}

How can we transform the meaning of a sentence such that the output remains fluent while the transformation process is maximally transparent? Some symbolic methods, e.g., those based on word replacement using taxonomy lookups \citep{bolshakov2004synonymous, huang2009using}, provide transparency and control but often produce unnatural sentences with limited variation. On the other end, of course, we now have the all-dominating paradigm of LLM prompting. But clearly this process is opaque, and any control is, at best, indirect \citep{greenblatt2024ai, 10.1162/tacl_a_00681}.

To bridge the gap between these paradigms, and to highlight an alternative to LLM-based methods, we propose \model, a neuro-symbolic framework for text manipulation. The process begins with a parser that maps a sentence onto a symbolic graph-based meaning representation, such as an Abstract Meaning Representation \citep[AMR,][]{banarescu2013abstract}. We posit that such a representation provides an effective interface for applying well-defined, precise, and targeted meaning-altering operations. Once the graph is updated, a generator converts it back into natural text. 

The process is illustrated in Figure \ref{fig:srl-switch}, where a single sentence is transformed into two new sentences with different meanings. On the left, we reverse semantic roles (e.g., \textit{tiger}, \textit{snake}), modifying the event structure of the sentence by swapping patient and agent roles. On the right, we apply a two-step process: removing the patient (\textit{tiger}) and negating the main predicate (\textit{bite-01}), resulting in a new sentence that describes a “harmless” snake. 
Not shown in Figure \ref{fig:srl-switch} but discussed later, we also introduce an optional faithfulness check as a post-processing step to evaluate the consistency between the transformed graph and the generated sentence, filtering out outputs affected by any eventual parsing or generation errors.

We argue that this combination of symbolic control, neural fluency, and post-hoc verification provides a powerful and flexible approach to sentence meaning transformation, offering a more controllable alternative or complement to LLMs. 

To illustrate the practical value of \model, we conduct a detailed demonstration study, leveraging our framework to generate hard negative pairs that pose significant challenges for text embedding models. By applying well-defined transformation rules targeting specific semantic phenomena, \model creates nuanced sentence pairs that go beyond superficial lexical differences, enabling a more fine-grained evaluation of embedding models. Through this application study, we shed light on their strengths and weaknesses in handling particular linguistic phenomena, addressing a limitation in current benchmarking practices, where such intricacies tend to be obscured.

Broadly speaking, this study contributes to the ongoing effort to pinpoint the limits of Transformer-based Neural Networks' understanding of sentence meaning. It has been established that even the latest LLMs struggle with important sentence-level meaning phenomena such as negation \citep{parmarLogicBenchSystematicEvaluation2024}, and \citet{xu2025large} finds that LLMs still struggle with exact, syntax-based modes of reasoning, sometimes being outperformed by BART \citep{lewis2019bart}, a much smaller and simpler sequence-to-sequence transformer. The fine-grained and dynamic experimental settings enabled by \model can shed new light on this central, but still open question regarding the abilities of Transformers.

The remainder of this paper is structured as follows: Section \S \ref{sec:rw} reviews background and related work. Section \S \ref{sec:ss} introduces our neuro-symbolic \model framework, detailing its components, including the parser, symbolic graph transformations, generator, and optional faithfulness checker. Section \S \ref{sec:app} presents an application demonstration, focusing on generating hard negative pairs to challenge state-of-the-art text embedding models and reveal potential weaknesses in their linguistic understanding. Finally, Section \S \ref{sec:concl} concludes the paper, discussing the broader implications of our work and potential directions for future research. We release code and data under a public license.\footnote{\url{https://github.com/impresso/sentence-smith}}

\section{Related Work and Background}
\label{sec:rw}

\paragraph{Method: Semantic graph as an intermediate representation.} The value of using semantic graphs as an intermediate representation, particularly AMR, has been highlighted in two recent surveys \citep{sadeddine-etal-2024-survey, wein-opitz-2024-survey}. This approach allows the fusion of neural network power with the expressivity and explicitness of meaning representations, which is especially useful when \textit{interpretability} and \textit{control} are required in an application. Related work applies this principle to style transfer \citep{jangra-etal-2022-star}, in the MT domain \citep{wein-schneider-2024-lost}, and data augmentation \citep{min-etal-2020-syntactic, shou-etal-2022-amr, shou-lin-2023-evaluate, ghosh-etal-2024-abex, kim2024simple}. Our work generalizes this approach further and imposes an additional check for validating the faithfulness of the generation; Unlike \citet{li-etal-2020-linguistically} we do not rely on an inflexible rule-parser and/or LLMs for verification, proving the feasibility of efficiently scaling this approach. Notably, the idea of planning/controlling sentence generation through meaning representation dates decades back \citep[i.a.,][]{sondheimer-nebel-1986-logical, mann1986systemic, kasper-1989-flexible, wijnen1990development, bateman1990interfacing}, but was limited by inaccuracies in parsing and generation. With stronger parsing and generation systems now available, we argue that effective usage has become feasible.

\paragraph{Application: Embedding models and benchmarking.} Text embedding models are crucial for a wide range of NLP tasks, including semantic search, information retrieval, and NLG evaluation \citep{clark2019sentence, muennighoff2022sgpt, gao2023retrieval}. Since \citet{reimers-gurevych-2019-sentence}'s foundational ``SBERT'' work, multiple branches of embedding model research have emerged. These include enhancing model performance through scaling parameters \citep{wang2023improving} or training data \citep{wang2022texte5}, as well as exploring unsupervised embeddings \citep{gao-etal-2021-simcse} and interpretable embeddings \citep{opitz-frank-2022-sbert}. Key questions are: \textit{What is the accuracy of such embeddings? \textbf{What level of linguistic understanding resides in those vectors?}} While the first question is typically addressed through large-scale benchmarks like MTEB \citep{muennighoff-etal-2023-mteb}, the second question requires a more fine-grained approach. Moreover, the questions are intertwined, and limits on linguistic understanding might be concealed by averages over large benchmark datasets where individual datasets often have hardly interpretable notions of similarity. We employ \model and demonstrate its capacity to generate fine-grained evaluation data that test embedding models' ability to assess different linguistic phenomena. With special regard to interpretability of text embeddings \citep{opitz2025interpretable}, our work is also related to a recent/concurrent line of work by Nastase et al.\ \citep{nastase-merlo-2024-tracking, nastase-etal-2024-exploring, nastase-etal-2024-exploring-syntactic}. In contrast to these works, however, we don't rely on costly human data creation.

Furthermore, the stasis of most benchmarks has also drawn criticism \citep{fan-etal-2024-nphardeval}, and limits evaluation to the available data, with potential ramifications for the trustworthiness of results. By demonstrating how \model can generate challenging, trustworthy, and interpretable test sets, we pave the way for more customizable, dynamic and interpretable testing of models.

\paragraph{Paraphrases and building minimal pairs.} How do two texts relate? In a broader context, our work is related to measuring paraphrases and entailment, both long-standing topics of interest in the NLP and ML \citep{bhagat2013paraphrase, bowman-etal-2015-large, zhou-bhat-2021-paraphrase, opitz-etal-2023-amr4nli, gubelmann-etal-2023-truth, NEURIPS2023_575c4500}. Instead of building or rating paraphrases, we first and foremost build challenging negatives given a paraphrase set, in a controlled manner, such that the relation between negative and paraphrase is transparent (e.g., negation). This may shed also more light on a problem recently highlighted: Notions of paraphrases empirically differ \citep{michail-etal-2025-paraphrasus} and the differences are not transparent. The controlled manipulations furnished by our \model framework allow for more in-depth studies. There are also other efforts that note the value of constructing such relation-controlled minimal linguistic pairs \citep[BLiMP,][]{warstadt-etal-2020-blimp-benchmark, jumelet2025multiblimp}---in contrast to these works, our framework does not require static resources or manual annotation effort.\footnote{Concurrent work also used LLMs to similarly generate hard negatives from given paraphrases, with similar application scenario \citep{michail2025examiningmultilingualembeddingmodels, magomere-etal-2025-claims}. However, this is a categorically different approach, because no controllable intermediate representation is leveraged, which limits the control over the particular type of relation. Also, the LLM use makes such approaches much more costly, while our approach only depends on smaller neural models.}

\section{The \model Framework}
\label{sec:ss}

\subsection{Overview}

The formal description of \model is straightforward. Given $s$ as an input sentence, a parser $p$ and a generator $g$, \model conducts a controlled transformation, resulting in a changed sentence $s'$, that is:
\begin{equation}
    s' =g \circ t_n \circ ... \circ t_1 \circ p ~|~ s,
\end{equation}
where $\{t_1, ..., t_n\}$ are graph transformations that we apply on the graph representation of the input sentence $s$, output of $p$. Finally, $g$ generates a text from the final state of the graph.

\subsection{Parameterization}

\paragraph{Graph model.} We rely on Abstract Meaning Representation \citep[AMR,][]{banarescu2013abstract} as our graph framework. AMR represents text as directed acyclic graphs, where nodes denote entities and edges capture semantic relations. The overarching goal of AMR is to explicitly encode ``Who does what to whom?'' This explicitness is a key motivation for our work, as it enables targeted meaning changes within the semantic structure.

\paragraph{Parsing and generation.} As is standard in most AMR applications, we employ off-the-shelf parsing and generation models.\footnote{\url{https://github.com/bjascob/amrlib}} These models, based on pre-trained BART \citep{lewis2019bart}, produce linearized sequence graphs from text (parsing) or text from linearized sequence graphs (generation). To facilitate manipulation, we convert the linearized graph into triples of the form \edge{u}{r}{v}, where \texttt{u} and \texttt{v} are graph nodes, and \texttt{r} is a relation. 

\paragraph{Graph transformation.} This is the interface where a human (or other controller) can influence the machine-based sentence transformation. Beyond various graph operations, we may wish to, for example, modify truth values by adding a negation to a select predicate, swap semantic roles, or add new information in a controlled manner. Since different use cases of \model may require distinct transformations, the specific rules applied in our demonstration study are detailed later in \S \ref{sec:app}. 

\paragraph{Validation.} Once a transformed sentence is generated, we would like to validate whether the intended transformation was successfully executed. Potential failure cases include noise introduced by the parsing and generation process. We employ a faithfulness check function $check(s, s') \in \{-1, 0, 1\}$, where $-1$ denotes contradiction between $s$ and $s'$, $0$ implies a neutral relation, and $1$ indicates that $s'$ is entailed by $s$. 

The criteria for discarding output sentences depend on the application: If the goal is meaning alteration, we should filter out cases where $s$ and $s'$ mutually entail each other. Conversely, if we seek paraphrases, we should discard cases where contradiction or neutrality is detected. \model parameterizes $check$ with an efficient NLI-based model from \citet{steen-etal-2023-little}. This system is robustness-enhanced through data augmentation and achieves strong results on the TRUE faithfulness benchmark \citep{honovich-etal-2022-true-evaluating}.\footnote{We use \url{https://huggingface.co/juliussteen/DeBERTa-v3-FaithAug}.}

\section{Application Study}
\label{sec:app}

As an important application to demonstrate the usefulness of \model, we focus on fine-grained linguistic testing of embedding models. 

\begin{figure}
\centering
\includegraphics[width=\linewidth]{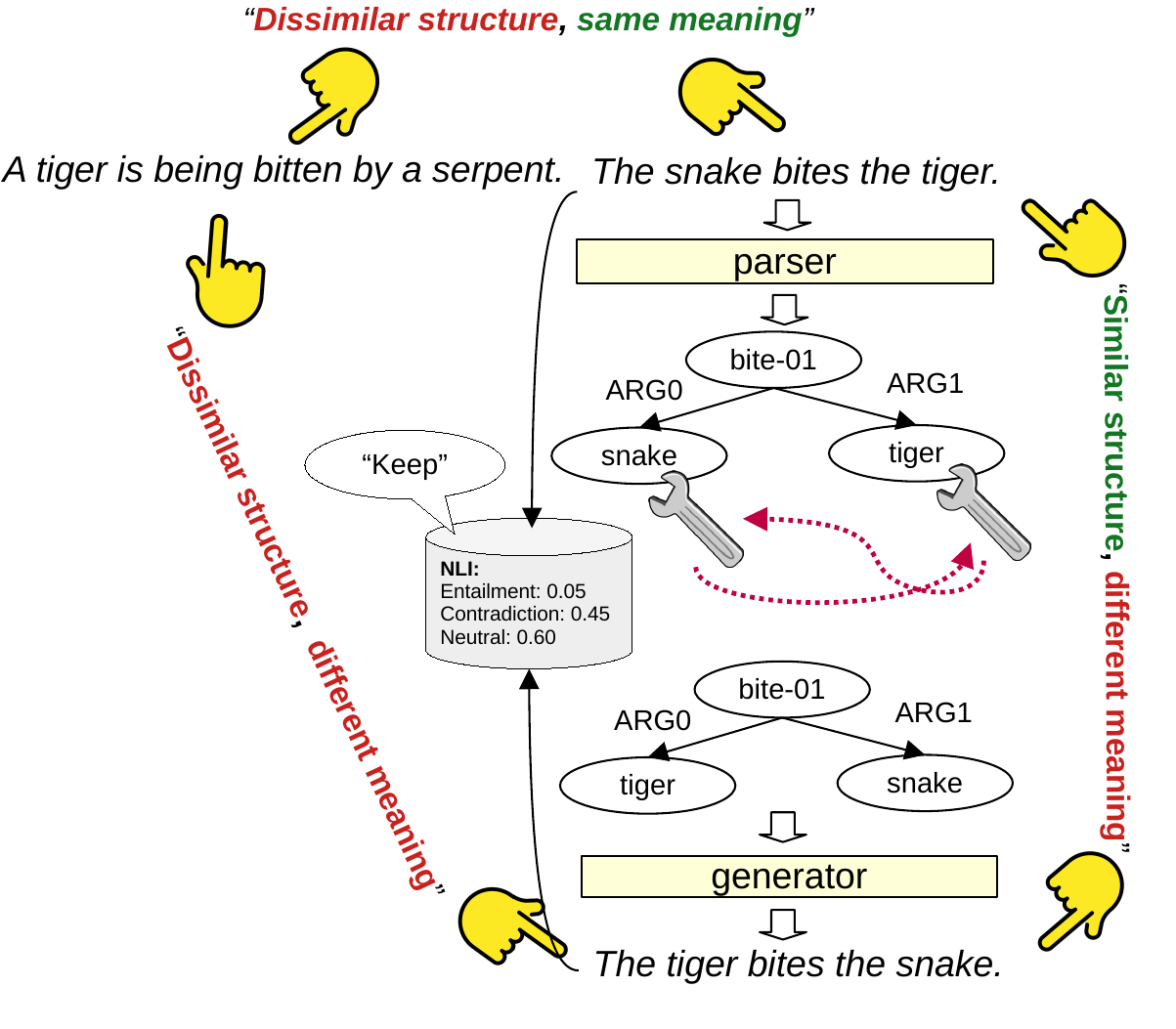}
\caption{Breaking the paraphrase relation and creating a challenging test case: The new sentence has a high surface similarity to the input, but it is not a paraphrase. By contrast, the actual paraphrases have a lower superficial overlap, thus posing a challenge to any model, especially those that tend to rely on surface cues.}
\label{fig:foilgen}
\end{figure}

\paragraph{Foil idea.} The foil concept is illustrated in Figure \ref{fig:foilgen}. The objective is to generate a new sentence that is structurally highly similar to an existing one but semantically distinct. If this transformation is applied to one part of a paraphrase pair, the resulting sentence will not maintain a paraphrase relation with either of the original sentences. In the example, we start with the paraphrases a)\textit{ A tiger is being bitten by a serpent.} and its paraphrastic variant b) \textit{The snake bites the tiger.} Our parser processes sentence b), and within the generated meaning graph, we perform a \textit{semantic role confusion}, swapping the agent and patient of the \textit{bite} event. Finally, our generator produces c) \textit{The tiger bites the snake}, which is structurally more similar to a) than a) and b) are to each other (i.e., higher lexical overlap). However, the event’s semantics have \textit{fundamentally} changed, thereby breaking the paraphrase relation.

\subsection{Setup}

\subsubsection{Defining Semantic Manipulations}

\begin{figure}
    \centering
    \includegraphics[width=\linewidth]{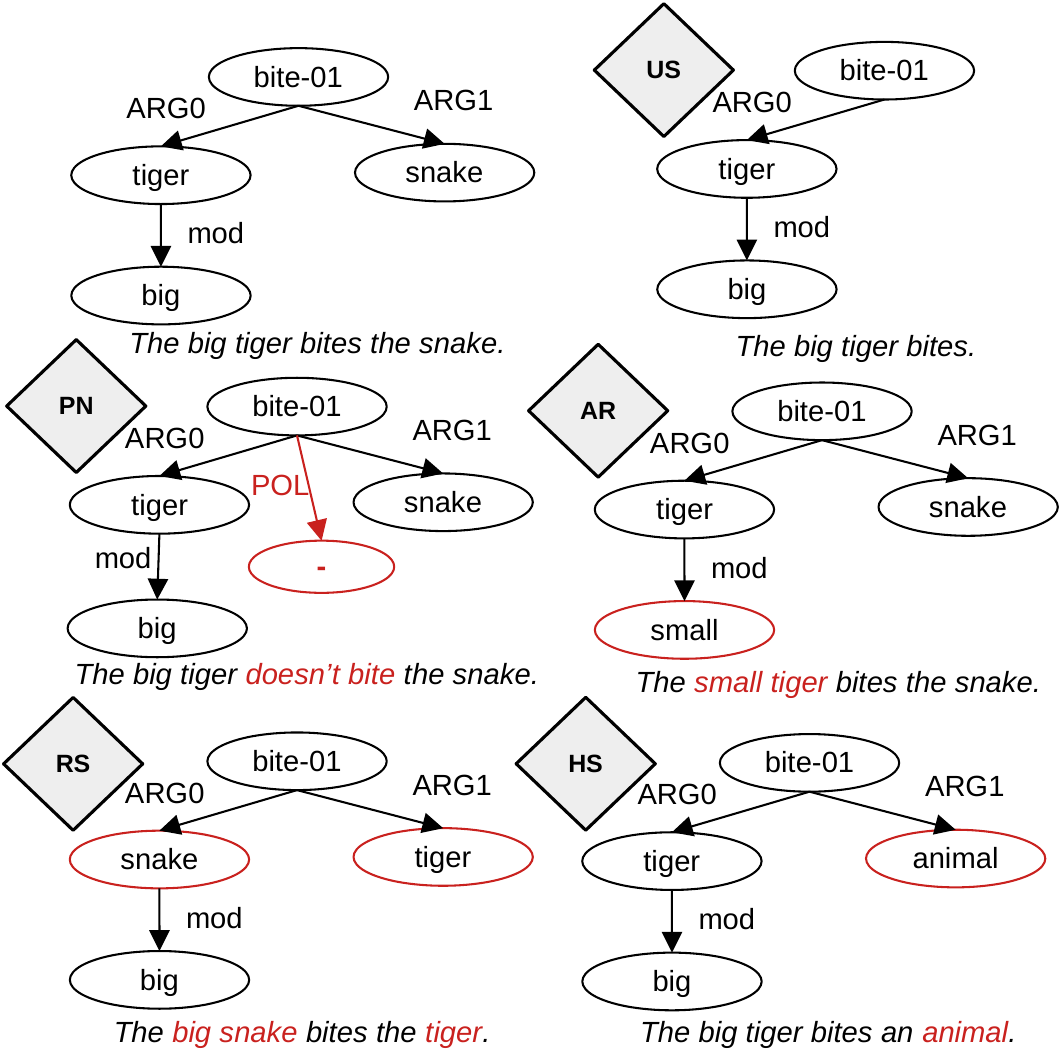}
    \caption{Five aspectual semantic manipulations.}
    \label{fig:five-rules}
\end{figure}

We define five semantic manipulations designed to generate foils from a given sentence: Polarity Negation, Role Swap, Underspecification, Antonym Replacement, and Hypernym Substitution. Examples of all five manipulation types are displayed in Figure \ref{fig:five-rules}. Below, we present a more detailed description of each manipulation.

\paragraph{PN: Polarity Negation.} This fundamental manipulation assesses embedding models' sensitivity to negation. We define negation as altering the polarity of a predicate within an AMR graph. Specifically, this transformation attaches a \edge{n}{:polarity}{-} edge (in AMR notation, a negation) to a randomly selected node \texttt{n}, excluding pronouns. For example, modifying the predicate \textit{approve} by adding this edge results in its opposite, \textit{disapprove} (i.e., in AMR: \edge{n}{:polarity}{-} is added to node \texttt{n}, where \edge{n}{:instance}{approve} already exists). The manipulated AMR is then used to generate the negated text. As can be seen in the corresponding example of Figure~\ref{fig:five-rules}, the truth value of the sentence changes.

\paragraph{RS: Role Swap.} This manipulation evaluates how well embedding models distinguish fine-grained semantic relations. It randomly swaps two graph nodes \texttt{u} and \texttt{v}, generalizing the semantic role switch illustrated in Figure \ref{fig:srl-switch} and Figure \ref{fig:five-rules}, where \texttt{s} (instance of snake) and \texttt{t} (instance of tiger) were swapped. 

\paragraph{US: Underspecification.} This transformation removes a randomly selected leaf node \texttt{n} from the AMR graph, reducing semantic content, thereby breaking a paraphrase relation. The introduced ambiguity or incompleteness  tests how robust embedding models are to partial information.\footnote{Note that it can also introduce a polarity change, in case a negation node in the AMR is removed. The notion of ``Underspecification'' refers to the AMR structure being reduced.} In Figure \ref{fig:five-rules}, the patient (a snake) has disappeared. 

\paragraph{AR: Antonym Replacement} inverts the meaning of a selected node while maintaining the context of the surrounding graph. The system identifies a random non-pronoun node, extracts its lexical stem (e.g., happy in happy-01), and queries WordNet \citep{miller-1993-wordnet} for antonyms. If a suitable antonym is found, the node label is updated while preserving any sense suffix (e.g., -01). This transformation alters meaning more explicitly than polarity negation by directly substituting an opposite term. In the Example for AR given in Figure~\ref{fig:five-rules}, same as in PN, the truth value changes ---while still a \textit{biting} is happening, this time the agent is small rather than big. 

\paragraph{HS: Hypernym Substitution.} This transformation replaces a word with its hypernym, distorting inferences and breaking paraphrase relationships. Consider \textit{Penguins can’t fly}. If \textit{Penguin} is replaced with its hypernym \textit{Bird}, the modified sentence is false (``Birds can’t fly''), demonstrating how hypernym substitution alters meaning while preserving surface structure. Following the same WordNet-based approach as in the  antonym replacement, non-pronoun node labels are substituted with randomly selected hypernyms. The node’s graph position and semantic roles remain unchanged, but abstraction increases, disrupting logical inferences. In Figure \ref{fig:five-rules}, we observe how \textit{snake} is replaced with \textit{animal}. This is also some form of underspecification, but speaking through the AMR, it is achieved by replacing (not removing) a concept (which can also cause a change in truth value, as in the aforementioned bird-example).

\subsubsection{Data Instantiation}

\paragraph{Initial datasets.} We use two paraphrase datasets as our base data. The first is PAWS \citep{yang-etal-2019-paws}\footnote{\url{https://huggingface.co/datasets/google-research-datasets/paws-x/tree/main}}, from which we take the positive pairs. While PAWS provides structurally similar paraphrases—originally designed to challenge paraphrase detection models with adversarial positive and negative pairs—it is not fully ideal for our purpose. Due to the high structural similarity, we anticipate slightly lower difficulty in the challenge pairs generated from this dataset.

To obtain a more comprehensive benchmark, we additionally leverage a more diverse set of ChatGPT-generated paraphrases \citep{chatgpt_paraphrases_dataset}\footnote{\url{https://huggingface.co/datasets/humarin/chatgpt-paraphrases}}, henceforth denoted as GPTP. This dataset consists solely of positive paraphrase pairs. Also unlike PAWS, GPTP exhibits greater structural variability, making our resulting challenge set inherently more difficult. \label{par:statvalidationdifference} We validate this statistically by computing bag-of-words similarity between paraphrase pairs in each dataset: PAWS yields an average similarity of 0.9, whereas GPTP has a much lower average similarity of 0.4 (see Appendix \ref{ssec:bowdivergence} for details). 

\paragraph{Data induction and filtering.} After employing our parser $parse$, a graph transformation $manip$, and the generator $gen$ on every pair, we end up with triples $(s, p, f)$, where $s$ is the original sentence, $p$ a paraphrase, and $f=gen(manip(parse(s)))$ our generated foil (structurally highly similar to $s$, but of another meaning). For post-processing, to maximize the integrity of the generated benchmark, we apply a harsh quality filtering process through our NLI-based validation module: We only retain those ``contradiction'' sentence pairs with a semantic relationship confidence score greater than 90\%. This ensures that the final dataset focuses on maximally validated paraphrase foils. However, we will later also experiment with a different filtering criterion. Final dataset statistics, including distribution of transformations, are shown in Table \ref{tab:datastats}. 

\begin{table}
    \centering
    \adjustbox{width=\linewidth}{\begin{tabular}{lrrrrrr}
    \toprule
         & total & PN \% & RS \% & US \% & AR \% & HS \% \\
         \midrule
       PAWS & 1,737 & 39.5  & 18.5  & 9.7  & 27.2  & 5.0  \\
       GPTP & 11,456 & 36.5  & 15.3  & 9.6  & 30.4  & 8.2  \\
       \bottomrule
    \end{tabular}}
    \caption{Final data set statistics.}
    \label{tab:datastats}
\end{table}

\paragraph{Human evaluation of generation.}\label{ssec:humaneval} We conduct a human annotation to assess the quality of our generations. Specifically, we are interested in two variables. First, \textit{Does the meaning of sentence B differ from that of sentence A (no, or yes)?} This lets us assess how faithfully the final output of \model has achieved our main goal: The manipulation of text meaning. Second, we wonder about the fluency of the final generation: \textit{Given sentence A, is the fluency of sentence B worse, about the same, or improved?} We task two annotators to annotate 100 randomly sampled pairs from each of PAWS and GPTP, so 200 in total. 

Table \ref{tab:humananno} presents the results. Our primary goal—meaning manipulation—is achieved to an overwhelming degree, with only one or two exceptions across both samples. While fluency tends to degrade on average, a notable portion of outputs remain equally fluent or even improve in fluency.

\begin{table}[]
    \centering
   \adjustbox{width=\linewidth}{ \begin{tabular}{lrr|rrrrrr}
         & \multicolumn{2}{c|}{Meaning}  & \multicolumn{6}{c}{Fluency}\\
         & \multicolumn{2}{c|}{accuracy}    & \multicolumn{2}{c}{worse} & \multicolumn{2}{c}{same} & \multicolumn{2}{c}{better} \\
             \cmidrule{2-9}
             & A1 & A2 & A1 & A2 & A1 & A2 & A1 & A2 \\
        \midrule
        PAWS & 98 & 99& 34 & 41 &  30 & 33 &  36 & 26 \\
        GPTP & 100 & 99 & 42 & 45 & 35 & 37 & 23 & 18   \\
        \bottomrule
    \end{tabular}}
    \caption{Manual assessment of the generation quality. All numbers are \%s. A1 and A2 are human annotators.}
    \label{tab:humananno}
\end{table}

Examples of our generated foils are shown in Appendix \ref{app:examples}, Table \ref{tab:foilexamples}. E.g., through adding a negative polarity to the node ``need'', the statement \textit{But you need to get somebody like Warren to do it} becomes \textit{But you don't need to get somebody like Warren to do it}, which is superficially highly similar, but has the opposite meaning of the original paraphrase \textit{You should find someone similar to Warren to handle it} that is structurally much more different on the surface.

\subsection{Evaluating Embedding Models}

\subsubsection{Embedding Model Selection}

With our benchmark fully set up, we  evaluate 29 off-the-shelf text embedding models. The selection aims to balance  performance and efficiency, comprising top-ranked models at the time of writing as well as other widely used baselines. 
These include  LaBSE \cite{feng-etal-2022-languagelabse}, SBERT \citep{reimers-gurevych-2019-sentence}, Jina \citep{gunther-etal-2023-jina}, and E5 \citep{wang2022texte5}. Additional models were drawn from the MTEB  leaderboard\footnote{\url{https://huggingface.co/spaces/mteb/leaderboard}}  to provide broader coverage of model types.

It is important to note that names like ``SBERT'' or ``E5'' represent techniques rather than specific model instances. Therefore, from here on, we reference specific models by their Hugging Face identifiers. For instance, \embedder{all-mpnet-base-v2} refers to a widely used embedding model that applies the SBERT training methodology to a self-supervised pre-trained MPNET model \citep{NEURIPS2020_c3a690bempnet}.

\subsubsection{Evaluation Measures} 

Consider any model $\mathcal{E}$ that maps a pair of texts to a real-valued ``similarity score'' (in our case we have text embedding models that construct the embeddings, and the cosine similarity builds the similarity). To assess the quality of such a model on our induced benchmark, we compute two key evaluation metrics. The first is \textbf{triplet accuracy $(TACC)$}, a simple and interpretable classification-oriented score that measures the ratio of cases where the embedding model remains robust to the foil.  We consider a set with triplets $T = \{(s_i, p_i, f_i)\}_{i=1}^n$, where $s_i$ is a text, $p_i$ its paraphrase, and $f_i$ our generated foil designed to mislead the model into assigning it a higher similarity score. Then 
\begin{equation*}
    TACC = \frac{1}{|T|}\sum_{(s, p, f) \in T}\mathcal{I}[\mathcal{E}(s,p) > \mathcal{E}(s, f)],
\end{equation*}
\noindent where $\mathcal{I}[c]$ returns 1 if the condition $c$ is true, and 0 else. As a more standard metric that is commonly used to evaluate tasks where a floating value (here: similarity) must be compared against a discrete label (here: paraphrase/not-paraphrase), we also measure the \textbf{Area Under the Receiver Operator Curve ($AUC$)} that represents a softer score directly based on the real valued output scores of embedding models. For this, we dichotomize the dataset into positive pairs $\{(s_i, p_i)\}_{i=1}^n$ and negative pairs $\{(s_i, f_i)\}_{i=1}^n$, essentially inducing a binary paraphrase classification task. Since it is hard to say which of the two metrics is generally more informative, models should clearly excel in both and thus \citep[Rec.\ 3 in Section 8]{10.1162/tacl_a_00675}, to compute a single ``performance'' number for a given dataset, we are using an harmonic mean of TACC and AUC. 

\subsection{Results}
\begin{table}[ht]
    \centering
    \adjustbox{width=\linewidth}{%
    \begin{tabular}{lrr|r}
        \toprule
        \textbf{Model Name} & \textbf{PAWS} & \textbf{GPTP} & \textbf{AVG} \\
        \midrule
\embedder{sentence-t5-large} & 0.9506 & 0.8026 & 0.8704 \\
\embedder{ember-v1} & 0.9562 & 0.7513 & 0.8415 \\
\embedder{bge-base-en-v1.5} & 0.9506 & 0.6910 & 0.8003 \\
\embedder{e5-base-v2} & 0.9378 & 0.6930 & 0.7970 \\
\embedder{GIST-Embedding-v0} & 0.9297 & 0.6954 & 0.7957 \\
\embedder{FAB-Ramy-v1} & 0.9190 & 0.6928 & 0.7900 \\
\embedder{gte-base-en-v1.5} & 0.9062 & 0.6947 & 0.7865 \\
\embedder{all-mpnet-base-v2} & 0.9046 & 0.6920 & 0.7841 \\
\embedder{instructor-base} & 0.9242 & 0.6503 & 0.7634 \\
\embedder{paraphrase-MiniLM-L12-v2} & 0.9509 & 0.6286 & 0.7569 \\
\embedder{nomic-embed-text-v1.5} & 0.8896 & 0.6519 & 0.7524 \\
\embedder{jina-embeddings-v2-base-en} & 0.9314 & 0.5965 & 0.7272 \\
\embedder{MedEmbed-small-v0.1} & 0.9237 & 0.5980 & 0.7260 \\
\embedder{stella-base-en-v2} & 0.9508 & 0.5724 & 0.7146 \\
\embedder{Wartortle} & 0.9556 & 0.5639 & 0.7093 \\
\embedder{LaBSE} & 0.9657 & 0.5444 & 0.6963 \\
\embedder{all-MiniLM-L12-v2} & 0.9234 & 0.5586 & 0.6961 \\
\embedder{gtr-t5-large} & 0.8501 & 0.5762 & 0.6869 \\
\embedder{cde-small-v1} & 0.8507 & 0.5653 & 0.6792 \\
\embedder{gte-micro} & 0.9320 & 0.5315 & 0.6769 \\
\embedder{msmarco-bert-co-condensor} & 0.8724 & 0.5474 & 0.6727 \\
\embedder{contriever-base-msmarco} & 0.8880 & 0.5214 & 0.6570 \\
\embedder{snowflake} & 0.8852 & 0.5104 & 0.6475 \\
\embedder{Ivysaur} & 0.9194 & 0.4889 & 0.6384 \\
\embedder{Venusaur} & 0.9323 & 0.4459 & 0.6033 \\
\embedder{distiluse-base-v2} & 0.9475 & 0.4393 & 0.6003 \\
\embedder{allenai-specter} & 0.8143 & 0.3908 & 0.5281 \\
\embedder{SGPT-125M} & 0.7491 & 0.3982 & 0.5200 \\
\embedder{komninos} & 0.8905 & 0.3431 & 0.4953 \\
        \bottomrule
    \end{tabular}}
    \caption{Main results. Shown numbers are harmonic means of AUC and TACC measures. AVG is their arithmetic mean across datasets.}
    \label{tab:mainres}
\end{table}

Table \ref{tab:mainres} shows the main results. The top-ranked model is \embedder{sentence-t5-large}, showing an AVG score of 0.87, outperforming the worst-ranked model \embedder{SGPT-125M} by about 30 percentage points (pp.). On PAWS, several models show strong performance of more than 0.90. This is likely due to the higher initial structural similarity of the paraphrases (for statistics, see the Appendix, \S \ref{app:otherstats}), reducing the effectiveness of some foils. On GPTP, with its more varied structural differences, the differences between models grows, and even the best performing model only barely exceeds an  AVG of 0.8. For individual metrics on each dataset, see Appendix \ref{app:otherstats}: Table \ref{tab:gpt} shows evaluation on GPTP, and Table \ref{tab:pawsx} on PAWS.

\subsection{Analysis}

\begin{figure*}[ht]
    \centering    \includegraphics[width=1.0\linewidth]{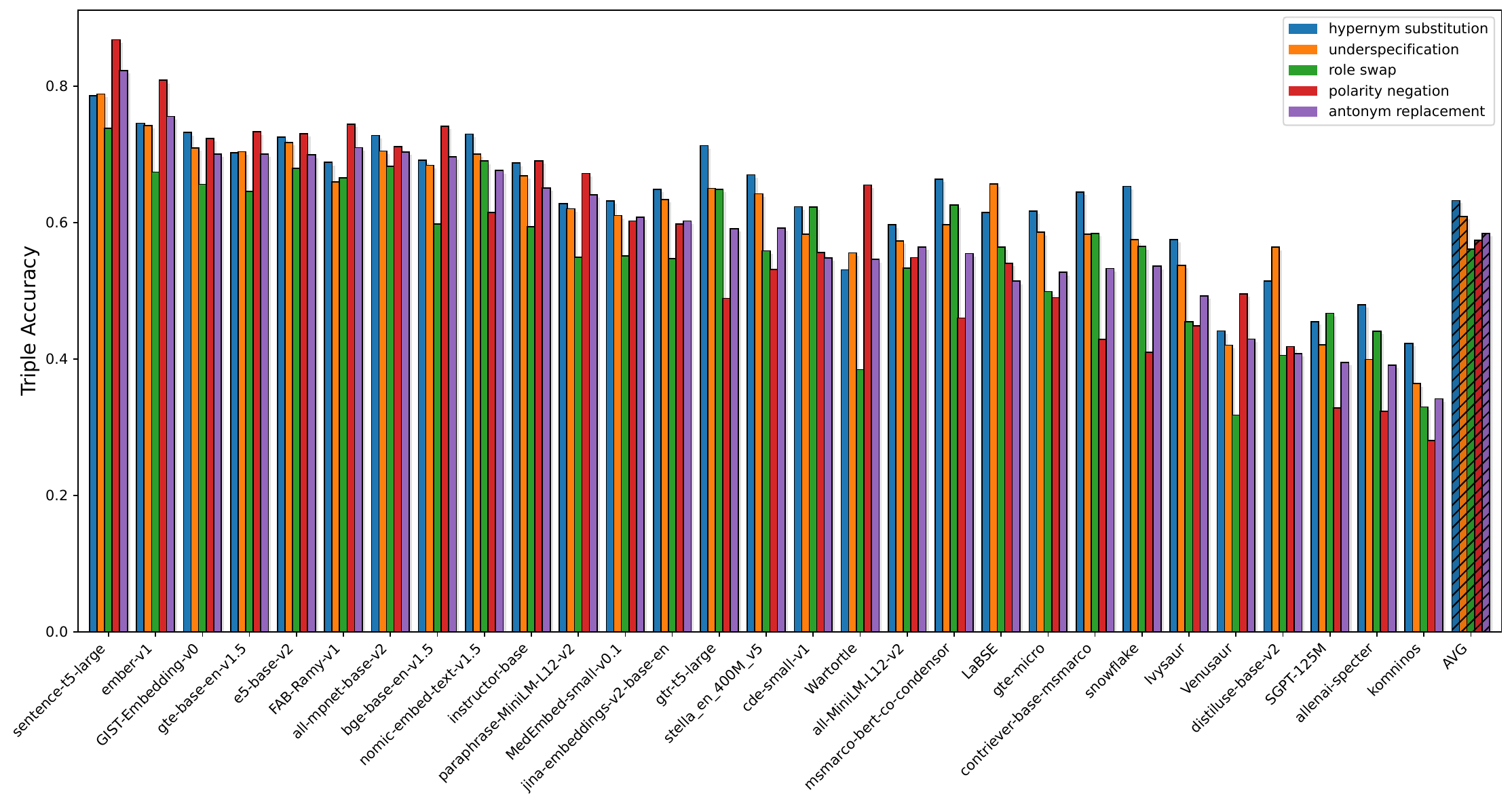}
    \caption{Fine-grained linguistic benchmarking of text embedding models.}
    \label{fig:fine-view-performance}
\end{figure*}

A fine-grained picture is shown in Figure \ref{fig:fine-view-performance}. For each  embedding model, we plot the $TACC$ accuracy across different manipulation types within the GPTP dataset. We make several interesting observations: 1.\ \embedder{sentence-t5-large} excels in all categories. 2.\ Many embedding models are moderately robust to polarity changes (purple and red bars), an important property for models of which we expect finer linguistic understanding. However, this is not consistent across models, with some exhibiting a $TACC$ accuracies at or below 40\%. Notably, the (less recent) \embedder{all-mpnet-base-v2} shows the most balanced performance across all manipulations, suggesting its suitability for use cases requiring uniform sensitivity to diverse linguistic manipulations. On average (AVG, right column), role swap is the most difficult transformation, while hypernym substitution is the easiest, though the differences are not pronounced. Overall, the results indicate considerable potential for improvement in embedding models’ fine-grained linguistic understanding.

\paragraph{NLI filtering ablation study.} To ensure the highest faithfulness of our automatically induced challenge cases, we selected only pairs where our NLI-based validation assigned a high probability of contradiction. In this experiment, we construct an alternative dataset containing more ambiguous cases by selecting pairs labeled as \textit{neutral} with probabilities between 50\% and 80\%. Specifically, we examine how much the final ranking deviates  between the two  datasets.

The results are shown in Table \ref{tab:filtered-gpt}, Appendix~\ref{app:otherstats}. Interestingly, \embedder{sentence-t5-large}, consistent with all previous experiments, again achieves the top rank, even though the intersection of the two datasets is zero. Overall,  the ranking stays mostly stable (Spearman's rank correlation coefficient: 0.91), with no substantial changes. 

\paragraph{Comparison against \textit{MTEB} ranking.} Our created testing challenge focuses on \textit{linguistic understanding} and can be dynamically generated and adapted. On the other hand, static and large benchmarks like MTEB cover a large spectrum of \textit{tasks}, including STS, but also IR, argument mining, and so on. Therefore, we might expect a different ranking of models. Models that would score high in our benchmark but lower on MTEB could point at unleveraged linguistic strengths of a model that might not be fully reflected by MTEB. On the other hand, a relatively lower ranking in our challenge may indicate a model weakness and suggest vulnerabilities in difficult cases. 

In our Appendix, Table~ \ref{tab:compare_ranking_mteb}, we compare the ranking of our models against the relative rankings on MTEB. Notably, the relatively best ranked model on MTEB (\embedder{cde-small-v1}, as of October 1st, 2024, the best model of a size smaller than 400M params) only reaches place 20 in our challenge, suggesting poorer semantic sentence understanding than \embedder{sentence-t5-large}, a model of similar size based on a seq-to-seq T5 transformer. Thus, our results suggest potential vulnerabilities of \embedder{cde-small-v1}, especially towards negation and polarity. To assess the overall difference between our ranking versus the MTEB ranking, we again calculate the Spearman's rank correlation. We receive a score of 0.54, suggesting a significant but only moderate alignment of the two benchmarks. While MTEB contains a large number of datasets, constituting a very large overall benchmark, our datasets have the advantage that they contain \textit{fresh, unseen benchmarking data} that can be dynamically generated, and that allow for targeted probing of models' \textit{linguistic understanding}.

\paragraph{Embeddings from very large models.} We additionally use our data to probe sentence understanding of text embeddings from two very large language models: E5 with MISTRAL LLM 7B as backbone \citep{wang2022texte5,Jiang2023Mistral7}, and GTE with QWEN 7B LLM as backbone \cite{li2023towards, bai2023qwen}. 
\begin{table}
    \centering
        \adjustbox{width=\linewidth}{
    \begin{tabular}{l|rrrrr}
    \toprule
Metric         & \textit{LOWER} & E5-MISTRAL & GTE-QWEN & \textit{UPPER}  \\
TACC  & 0.326 & 0.416 & 0.540 & 0.820 \\
\bottomrule
    \end{tabular}}
    \caption{Results for embeddings from very large models on GPTP dataset. \textit{LOWER} and \textit{UPPER} are reference values from the other models tested on GPTP. The worst performing being \embedder{komnios} and the best performing \texttt{sentence-t5-large}.}
    \label{tab:large}
\end{table}
Interestingly, the resulting scores (Table~\ref{tab:large}) are surprisingly low, placing the models in the lower-middle rankings.\footnote{The weakest aspect is polarity, where both models get more than 50\% of cases wrong.}. In contrast, smaller models, specifically T5-based embeddings, perform better. Our hypothesis is that, since these LLM-based models are optimized for retrieval and longer-context tasks, their  embeddings may be less sensitive to fine-grained aspects of  sentence meaning.

\section{Conclusion}
\label{sec:concl}

Our \model framework facilitates controlled and highly transparent meaning transformations of text. We showcased \model's utility in an NLP task: Dynamic benchmarking and analysis of text embedding models. Concretely, we used it to produce challenging foils from paraphrase pairs, carefully breaking the paraphrase relation while a high degree of superficial similarity. Through this, we shed light on the robustness of text embedding models and their ability to distinguish fine-grained linguistic phenomena. 

\section*{Limitations}

We note two main types of limitations: those that relate to the model of meaning representation, and those that affect conclusions drawn from our automatically induced benchmark.

\paragraph{Meaning Representation Model.} It is  important to note that \model directly scales with improvements in the area of meaning representations. While AMR has the great advantage of large data sets as well as reasonably robust parsers and generators, it also comes with limitations and drawbacks that have been, over the years, carefully outlined by research. There is the mono-linguality \citep{banarescu2013abstract}, then there may be meaning non-isomorphisms due to ambiguity \citep{wein-2025-ambiguity}, and lack of some scope \citep{pustejovsky-etal-2019-modeling} and tense aspects \citep{donatelli-etal-2018-annotation}. Moreover, while AMR parsers and generators seem to score high on benchmarks \citep[e.g.,][]{vasylenko-etal-2023-incorporating}, these tasks remain far from solved  \citep{manning-etal-2020-human, opitz-frank-2022-better, yang-schneider-2024-relative}, as is also supported by our output investigation where we observed some fluency issues, that may both be due to generation or parsing issues. This setup restricts our framework to English, because robust parsers and generators are not yet widely available for other languages. However,  in the context of multilinguality, promising advances in cross-lingual parsing have emerged through ``Universal Representations'' \citep{van2021designing}. Exploiting the wealth of AMR tools, we might also effectively use an off-the-shelf MT model to wrap the English AMR parsing and generation process \citep{uhrig-etal-2021-translate}, or use cross-lingual AMR parsing for certain languages \citep{vanroy-van-de-cruys-2024-less, kang-etal-2024-cross-lingual}. Given sufficient quality of the MT system, this simple approach could already work for some cross-lingual use-cases of our \model. 

\paragraph{On drawing conclusions about embedding model performance.} Embedding models are widely used in research and industry, making it important to investigate differences in their performance. Our application study and its induced dataset \textit{do} allow us to test models' sensitivity to paraphrases while also highlighting the specific linguistic phenomena that drive this sensitivity. However, different embedding objectives and downstream settings may prioritize relevance as the primary goal, in which case this testing strategy may not provide the most informative signal for model selection. The ranking of models in our setting is conditional on their ``similarity'' scores serving as proxies for semantic similarity, rather than on a neutral or universal evaluation of  embedding quality. Thus, the results should not be misinterpreted as a general quality difference between embedding models. Instead, they reflect the constraint that performance (or, more cautiously, behavior) is evaluated only with respect to specific linguistic paraphrase phenomena. A strength of this targeted evaluation, however, is that it enables us to diagnose model weaknesses in particular categories. Future work may adapt the manipulation rules in ways so that additional objectives can be effectively evaluated. 

\section*{Acknowledgments}

We thank the anonymous reviewers for their comments, and the meta-reviewer for their detailed feedback. Three authors received funding through the project \textit{Impresso – Media Monitoring of the Past II. Beyond Borders: Connecting Historical Newspapers and Radio}. Impresso is a research project funded by the Swiss National Science Foundation (SNSF 213585) and the Luxembourg National Research Fund (17498891).


\bibliography{custom}

\appendix

\section{Appendix}
\label{sec:appendix}

\subsection{Example of Data Generation Process}
\label{app:running_example}
\begin{figure*}
    \centering
    \includegraphics[width=\linewidth]{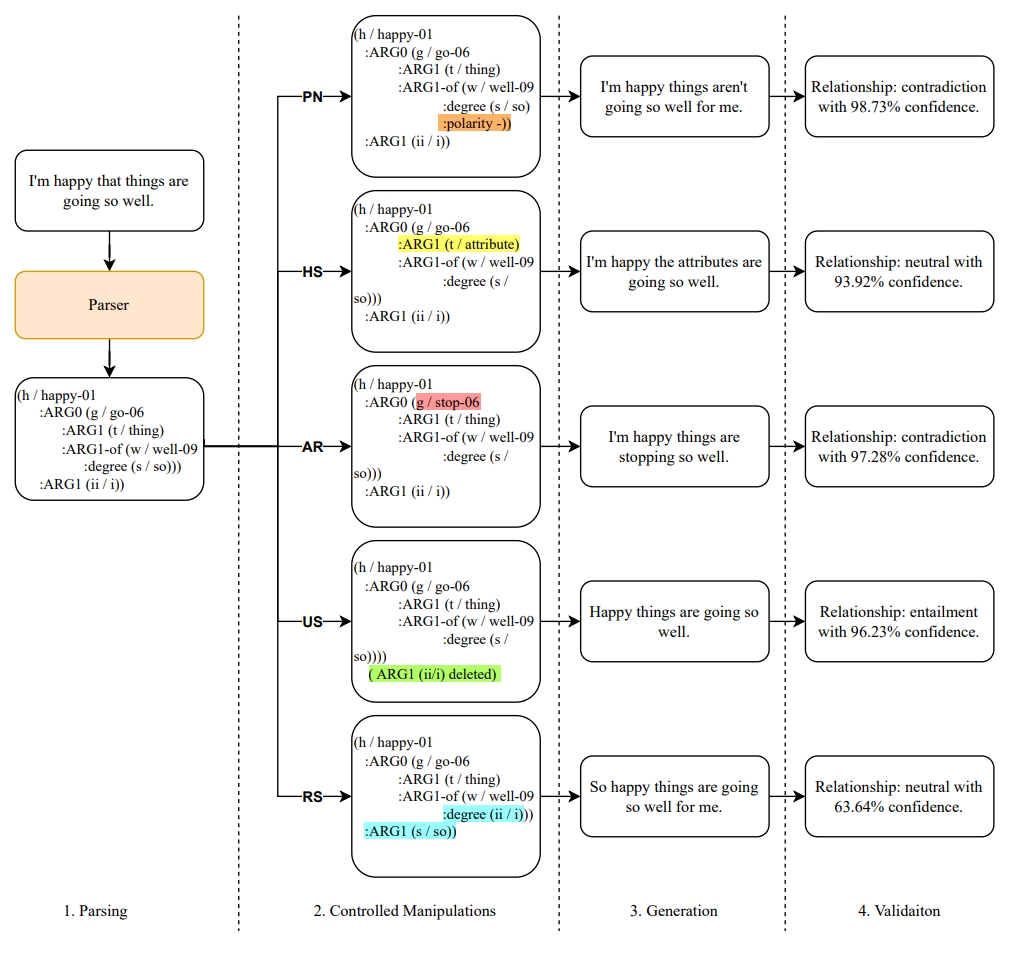}
    \caption{Running example of the full data generation process with \model. Under the strict quality filtering criterion of $>$ 90\% contradiction, only PN and AR are retained from the five candidate generations.  In the ablation study with the alternative, less stringent neutral filter, only the RS is retained.}
    \label{fig:manipulations}
\end{figure*}

A running example of a complete data generation and selection process with \model is shown in Figure~\ref{fig:manipulations}.

\textbf{PN}: the original sentence \textit{I'm happy that things are going so well} transforms into \textit{I'm happy things aren't going so well for me}. Here, we added \edge{w}{:polarity}{-} to node \texttt{w}, where \edge{w}{:instance}{well-09} represents ``well-being.'' While the resulting text may sound counterintuitive (``Happy things are not going well''), the goal is successfully achieved: the meaning is strictly altered (paraphrase relation is broken), while the surface structure remains highly similar with many overlapping tokens.

\textbf{RS}:  In our example, \textit{I'm happy that things are going so well} becomes \textit{So happy things are going so well for me}. In the AMR graph, \texttt{i} (instance of ``I'') and \texttt{s} (instance of ``so'') were swapped. While the original sentence implies an unspecified referent (\textit{who} is it that things are going well for?), the transformed sentence explicitly assigns this role to an entity (``me'') and also shifts emphasis (``so happy'' vs. ``so well''), altering meaning while preserving structural similarity.

\textbf{US}: \textit{I’m happy that things are going so well} is transformed into \textit{Happy things are going so well}. Here, the first-person subject (``I'') is removed, which could also lead to an alternative interpretation: that things themselves are happy. Arguably, this change could still be considered a paraphrase in some pragmatic contexts, and we would thus like to filter it out from our final data. This is exactly where the NLI validation module proves valuable—it assigns an entailment probability of 96.2\%, indicating that this particular instance might not be suitable as a foil.

\textbf{AR}: \textit{I’m happy that things are going so well} is transformed into \textit{I’m happy things are stopping so well}, where the predicate go-01 is replaced with stop-01 in the AMR graph, that is, \edge{g}{:instance}{go-01} becomes \edge{g}{:instance}{stop-01}. This results in a fundamental semantic shift: the original sentence conveys progression (things are going), while the transformed (slightly weird) sentence implies cessation (things are stopping), disrupting the paraphrase relation but maintains high lexical overlap.

\textbf{HS}: \textit{I’m happy that things are going so well} becomes\textit{I’m happy the attributes are going so well}. Here,``thing'', an already quite abstract term, is replaced with an even more abstract alternative, ``attribute'', leading to a grammatically valid but semantically odd sentence. In AMR, this corresponds to the change \edge{t}{:instance}{thing} $\rightarrow$ \edge{t}{:instance}{attribute}, demonstrating how even subtle lexical shifts can significantly impact meaning. The NLI validator assigns it a ``neutral''-label, with very high confidence. Thus, if wished for high strictness, this concrete example can also be filtered out.

\subsection{Bag-of-words Divergence Analysis}
\label{ssec:bowdivergence}
To further investigate the semantic divergence between the original sentences and their paraphrases in both datasets, we employ a simple bag-of-words approach. Specifically, we convert each sentence to lowercase, remove punctuation, split whitespace, optionally filter out English stopwords, and treat the remaining tokens as sets. We compute the \emph{Jaccard similarity} between the original and paraphrased token sets, defined as
\[
J(A, B) = \frac{|A \cap B|}{|A \cup B|},
\]
and use \(1 - J(A,B)\) as a simple measure of divergence. We compare two datasets: one from ChatGPT and one from PAWS. For the ChatGPT dataset, which consists of 11,457 sentence-paraphrase pairs, the average Jaccard similarity is approximately 0.3971. In contrast, the PAWS dataset (1,738 pairs) exhibits a much higher average Jaccard similarity of about 0.9002, indicating a much smaller structural divergence that leads to less challenging foils. This indicates that the ChatGPT paraphrases are, on average, much more lexically varied from the original sentences than those in PAWS. From a practical standpoint, this suggests that the ChatGPT-based paraphrasing process produces more diverse rewordings, potentially offering broader coverage for downstream tasks.

\subsection{Examples of Manipulations}
\label{app:examples}

Examples of some of our generated foils are shown in Table \ref{tab:foilexamples}.

\begin{table*}[t]
  \footnotesize
  \begin{tabular}{
    >{\raggedright\arraybackslash}p{0.4\linewidth}  
    >{\raggedright\arraybackslash}p{0.5\linewidth}  
    >{\centering\arraybackslash}p{0.05\linewidth}}   
  \toprule
  \textbf{Text} & \textbf{Choices} & \textbf{Type} \\
  \midrule

  The majority of Havocs served with the Soviets, but the US and Great Britain also used the planes.
  & \textbf{T:} Most Havocs were utilized by the Soviets, although the US and Great Britain also employed them.%
    \par\noindent\textbf{F:} Havocs also serve with the Soviet Union but the majority of their use is by the US and Britain.
  & RS \\
  \midrule

  Universities minister David Willetts said all universities can do is ask students if they have booked a flight home.
  & \textbf{T:} According to David Willetts, universities can only inquire about students' flight reservations for their return home.%
    \par\noindent\textbf{F:} University Minister David Blankts said the university could no longer do the asking whether students had booked flights home.
  & AR \\
  \midrule

  It was known as a relatively easy plane to fly with good handling during takeoff and landing.
  & \textbf{T:} The aircraft was recognized for its ease of operation and smooth handling during takeoff and landing.%
    \par\noindent\textbf{F:} It is known that it is not a relatively easy flying plane that is well-handled when taking off and landing.
  & PN \\
  \midrule

  The reason for the crash is not known.
  & \textbf{T:} The cause of the accident is uncertain.%
    \par\noindent\textbf{F:} The reason for the crash was known.
  & US \\
  \midrule

  Carcasses were also found near the contaminated watering holes.
  & \textbf{T:} The polluted watering holes were also discovered to have nearby carcasses.%
    \par\noindent\textbf{F:} The carcass was also found far from the contaminated water hole.
  & AR \\
  \midrule

  The auction house also sold one of the two gun belts owned by Jesse James at the time of his death.
  & \textbf{T:} One of Jesse James' two gun belts at the time of his death was also sold by the auction house.%
    \par\noindent\textbf{F:} The auction house also gave up one of Jesse James's two gun belts when he died.
  & HS \\
  \midrule

  Jean-Marc Wenger, who lives in Klingau, found the gold.
  & \textbf{T:} The gold was discovered by Jean-Marc Wenger, a resident of Klingau.%
    \par\noindent\textbf{F:} Jean-Marc Klonau a Wenger resident found gold.
  & RS \\
  \midrule

  Jean-Marc Wenger, who lives in Klingau, found the gold.
  & \textbf{T:} The gold was discovered by Jean-Marc Wenger, a resident of Klingau.%
    \par\noindent\textbf{F:} Jean-Marc Wenger who lives in Klanau has lost gold.
  & AR \\
  \midrule

  But at the moment it is a complete mystery.
  & \textbf{T:} Currently, it remains an enigma.%
    \par\noindent\textbf{F:} But it is not a complete mystery at the moment.
  & PN \\
  \midrule

  Thibaut Courtois was out quickly to thwart Sterling as Liverpool looked to get back in the game.
  & \textbf{T:} Liverpool attempted to make a comeback, but Thibaut Courtois swiftly prevented Sterling from scoring.%
    \par\noindent\textbf{F:} When Liverpool looked to get back in the game Thibaut Sterling was quick out to thwart Courttois.
  & RS \\
  \midrule

  Willis visited the Neon Museum in 2013 to celebrate her 90th birthday.
  & \textbf{T:} In 2013, Willis marked her 90th birthday by visiting the Neon Museum.%
    \par\noindent\textbf{F:} In 2013 to celebrate his 90th birthday he visited the Neo Museum.
  & US \\
  \midrule

  At least 3,000 Brussels bureaucrats earn more than David Cameron, it emerged yesterday.
  & \textbf{T:} Yesterday, it was revealed that over 3,000 Brussels bureaucrats earn a higher salary than David Cameron.%
    \par\noindent\textbf{F:} It emerged yesterday that at least 3000 bureaucrats in Brussels earn less than David Cameron.
  & AR \\
  \midrule

  We were going in water until hit the hill and spun.
  & \textbf{T:} We were traveling through water until we hit the hill and spun.%
    \par\noindent\textbf{F:} We went up the hill until the water hit and we spun.
  & RS \\
  \midrule

  But you need to get somebody like Warren to do it.
  & \textbf{T:} You should find someone similar to Warren to handle it.%
    \par\noindent\textbf{F:} But you don't need to get somebody like Warren to do it.
  & PN \\
  \midrule

  Julian E. Zelizer says Democrats should be questioning themselves on several key points.
  & \textbf{T:} Democrats ought to be reflecting on various crucial aspects, according to Julian E. Zelizer.%
    \par\noindent\textbf{F:} Julian E. Zelizer said Democrats should question him on several key points.
  & RS \\
  \bottomrule
  \end{tabular}
  \caption{Example cases from our automatically induced challenge benchmark. 
  T: The actual paraphrase. F: A generated foil, close to the input text in surface form, but different in meaning.}
  \label{tab:foilexamples}
\end{table*}

\subsection{Additional Tables}
\label{app:otherstats}

Table \ref{tab:gpt} presents embedding model rankings on the GPTP dataset, whereas Table \ref{tab:pawsx} reports rankings on the PAWS dataset. These two tables therefore correspond to different underlying data sources. In contrast, Table \ref{tab:filtered-gpt} provides additional results for GPTP only, obtained under an alternative quality filtering criterion in our NLI validator that retains more subtle examples with a probability between 50\% and 80\% of being labeled neutral.

\begin{table}[ht]
    \centering
    \adjustbox{width=\linewidth}{%
    \begin{tabular}{lrrr}
        \toprule
        \textbf{Model Name} & \textbf{TACC} & \textbf{AUC} & \textbf{AVG} \\
        \midrule
\embedder{sentence-t5-large} & 0.8197 & 0.7862 & 0.8026 \\
\embedder{ember-v1} & 0.7600 & 0.7427 & 0.7513 \\
\embedder{GIST-Embedding-v0} & 0.7050 & 0.6860 & 0.6954 \\
\embedder{gte-base-en-v1.5} & 0.7041 & 0.6856 & 0.6947 \\
\embedder{e5-base-v2} & 0.7111 & 0.6758 & 0.6930 \\
\embedder{FAB-Ramy-v1} & 0.7086 & 0.6776 & 0.6928 \\
\embedder{all-mpnet-base-v2} & 0.7049 & 0.6795 & 0.6920 \\
\embedder{bge-base-en-v1.5} & 0.6957 & 0.6864 & 0.6910 \\
\embedder{nomic-embed-text-v1.5} & 0.6622 & 0.6420 & 0.6519 \\
\embedder{instructor-base} & 0.6609 & 0.6401 & 0.6503 \\
\embedder{paraphrase-MiniLM-v2} & 0.6347 & 0.6227 & 0.6286 \\
\embedder{MedEmbed-small-v0.1} & 0.5991 & 0.5969 & 0.5980 \\
\embedder{jina-embeddings-v2-base-en} & 0.5988 & 0.5943 & 0.5965 \\
\embedder{gtr-t5-large} & 0.5780 & 0.5744 & 0.5762 \\
\embedder{stella-base-en-v2} & 0.5756 & 0.5693 & 0.5724 \\
\embedder{cde-small-v1} & 0.5717 & 0.5590 & 0.5653 \\
\embedder{Wartortle} & 0.5604 & 0.5674 & 0.5639 \\
\embedder{all-MiniLM-L12-v2} & 0.5570 & 0.5603 & 0.5586 \\
\embedder{msmarco-bert-co-condensor} & 0.5437 & 0.5511 & 0.5474 \\
\embedder{LaBSE} & 0.5529 & 0.5362 & 0.5444 \\
\embedder{gte-micro} & 0.5257 & 0.5375 & 0.5315 \\
\embedder{contriever-base-msmarco} & 0.5163 & 0.5265 & 0.5214 \\
\embedder{snowflake} & 0.5074 & 0.5135 & 0.5104 \\
\embedder{Ivysaur} & 0.4815 & 0.4966 & 0.4889 \\
\embedder{Venusaur} & 0.4362 & 0.4561 & 0.4459 \\
\embedder{distiluse-base-v2} & 0.4348 & 0.4439 & 0.4393 \\
\embedder{SGPT-125M} & 0.3889 & 0.4079 & 0.3982 \\
\embedder{allenai-specter} & 0.3818 & 0.4003 & 0.3908 \\
\embedder{komninos} & 0.3263 & 0.3617 & 0.3431 \\
        \bottomrule
    \end{tabular}}
    \caption[GPT Table]{Performance of models on GPTP dataset.}
    \label{tab:gpt}
\end{table}

\begin{table}[ht]
    \centering
    \adjustbox{width=\linewidth}{%
    \begin{tabular}{lrrr}
        \toprule
        \textbf{Model Name} & \textbf{TACC} & \textbf{AUC} & \textbf{AVG} \\
        \midrule
\embedder{LaBSE} & 0.9730 & 0.9586 & 0.9657 \\
\embedder{ember-v1} & 0.9597 & 0.9527 & 0.9562 \\
\embedder{Wartortle} & 0.9591 & 0.9521 & 0.9556 \\
\embedder{paraphrase-MiniLM-v2} & 0.9563 & 0.9455 & 0.9509 \\
\embedder{stella-base-en-v2} & 0.9534 & 0.9482 & 0.9508 \\
\embedder{sentence-t5-large} & 0.9551 & 0.9462 & 0.9506 \\
\embedder{bge-base-en-v1.5} & 0.9545 & 0.9467 & 0.9506 \\
\embedder{distiluse-base-v2} & 0.9522 & 0.9429 & 0.9475 \\
\embedder{e5-base-v2} & 0.9419 & 0.9337 & 0.9378 \\
\embedder{Venusaur} & 0.9396 & 0.9251 & 0.9323 \\
\embedder{gte-micro} & 0.9356 & 0.9284 & 0.9320 \\
\embedder{jina-embeddings-v2-base-en} & 0.9356 & 0.9273 & 0.9314 \\
\embedder{GIST-Embedding-v0} & 0.9361 & 0.9233 & 0.9297 \\
\embedder{instructor-base} & 0.9287 & 0.9198 & 0.9242 \\
\embedder{MedEmbed-small-v0.1} & 0.9321 & 0.9155 & 0.9237 \\
\embedder{all-MiniLM-L12-v2} & 0.9315 & 0.9155 & 0.9234 \\
\embedder{Ivysaur} & 0.9252 & 0.9137 & 0.9194 \\
\embedder{FAB-Ramy-v1} & 0.9298 & 0.9084 & 0.9190 \\
\embedder{gte-base-en-v1.5} & 0.9114 & 0.9010 & 0.9062 \\
\embedder{all-mpnet-base-v2} & 0.9131 & 0.8963 & 0.9046 \\
\embedder{komninos} & 0.9143 & 0.8680 & 0.8905 \\
\embedder{nomic-embed-text-v1.5} & 0.8982 & 0.8812 & 0.8896 \\
\embedder{contriever-base-msmarco} & 0.8964 & 0.8797 & 0.8880 \\
\embedder{snowflake} & 0.8930 & 0.8775 & 0.8852 \\
\embedder{msmarco-bert-co-condensor} & 0.8774 & 0.8674 & 0.8724 \\
\embedder{cde-small-v1} & 0.8843 & 0.8196 & 0.8507 \\
\embedder{gtr-t5-large} & 0.8619 & 0.8386 & 0.8501 \\
\embedder{allenai-specter} & 0.8228 & 0.8060 & 0.8143 \\
\embedder{SGPT-125M} & 0.7664 & 0.7326 & 0.7491 \\
        \bottomrule
    \end{tabular}}
    \caption[PAWS Table]{Performance of models on our PAWS dataset.}
    \label{tab:pawsx}
\end{table}

\begin{table}[ht]
    \centering
    \adjustbox{width=\linewidth}{%
    \begin{tabular}{lrrrrr}
        \toprule
        \textbf{Model Name} & \textbf{TACC} & \textbf{AUC} & \textbf{AVG} & \textbf{RANK} & \textbf{GROUP} \\
        \midrule
\embedder{sentence-t5-large} & 0.7260 & 0.6945 & 0.7099 & 1 & 1 \\
\embedder{FAB-Ramy-v1} & 0.6939 & 0.6638 & 0.6785 & 6& 1\\
\embedder{all-mpnet-base-v2} & 0.6691 & 0.6463 & 0.6575 & 7& 1 \\
\embedder{ember-v1} & 0.6635 & 0.6505 & 0.6569 & 2 & 1\\
\embedder{nomic-embed-text-v1.5} & 0.6454 & 0.6239 & 0.6345 & 9& 1 \\
\embedder{e5-base-v2} & 0.6449 & 0.6198 & 0.6321 & 5& 1\\
\embedder{GIST-Embedding-v0} & 0.6375 & 0.6238 & 0.6306 & 3& 1 \\
\embedder{gte-base-en-v1.5} & 0.6218 & 0.6144 & 0.6181 & 4& 1 \\
\embedder{bge-base-en-v1.5} & 0.6156 & 0.6007 & 0.6081 & 8& 1\\
\embedder{instructor-base} & 0.6009 & 0.5853 & 0.5930 & 10& 1 \\
\embedder{gtr-t5-large} & 0.5975 & 0.5795 & 0.5884 & 14 & 1 \\
\embedder{stella-base-en-v2} & 0.5862 & 0.5769 & 0.5815 & 15& 1 \\
\embedder{cde-small-v1} & 0.5795 & 0.5622 & 0.5707 & 16& 2 \\
\embedder{msmarco-bert-co-condensor} & 0.5609 & 0.5644 & 0.5626 & 19& 2 \\
\embedder{MedEmbed-small-v0.1} & 0.5490 & 0.5488 & 0.5489 & 12& 1 \\
\embedder{jina-embeddings-v2-base-en} & 0.5445 & 0.5500 & 0.5472 & 13 & 1\\
\embedder{snowflake} & 0.5479 & 0.5416 & 0.5447 & 23& 2 \\
\embedder{paraphrase-MiniLM-L12-v2} & 0.5457 & 0.5390 & 0.5423 & 11 & 1\\
\embedder{contriever-base-msmarco} & 0.5299 & 0.5346 & 0.5322 & 22& 2 \\
\embedder{LaBSE} & 0.5389 & 0.5211 & 0.5299 & 20& 2 \\
\embedder{all-MiniLM-L12-v2} & 0.5287 & 0.5310 & 0.5298 & 18& 2 \\
\embedder{gte-micro} & 0.4543 & 0.4702 & 0.4621 & 21& 2 \\
\embedder{Ivysaur} & 0.4476 & 0.4674 & 0.4573 & 24& 2 \\
\embedder{allenai-specter} & 0.4510 & 0.4532 & 0.4521 & 28& 2 \\
\embedder{Wartortle} & 0.4391 & 0.4411 & 0.4401 & 17& 2 \\
\embedder{distiluse-base-v2} & 0.4397 & 0.4377 & 0.4387 & 26& 2 \\
\embedder{SGPT-125M} & 0.4036 & 0.4219 & 0.4125 & 27& 2 \\
\embedder{komninos} & 0.3596 & 0.4050 & 0.3810 & 29& 2\\
\embedder{Venusaur} & 0.3579 & 0.3914 & 0.3739 & 25& 2 \\
        \bottomrule
    \end{tabular}}
    \caption[Filtered GPT Dataset Results]{Performance of models on GPTP dataset, when data for examples that are in between 50\% and 80\% neutral. RANK: The rank of the model when GPTP has been filtered by our main criterion (contradiction), compare with Table \ref{tab:gpt}. A more coarse view is the GROUP, it shows the binary (better, worse) group that a model is assigned to in the main data (again, c.f., Table \ref{tab:gpt}).}
    \label{tab:filtered-gpt}
\end{table}

\begin{table}[]
    \centering
    \adjustbox{width=\linewidth}{\begin{tabular}{lrr}
    \toprule
         & \textbf{RANK} & \textbf{MTEB rank (relative)} \\
         \midrule
\embedder{sentence-t5-large} & 1 & 16 \\
\embedder{ember-v1} & 2 & 5  \\
\embedder{bge-base-en-v1.5} & 3 & 4  \\
\embedder{e5-base-v2} & 4 & 9  \\
\embedder{GIST-Embedding-v0} & 5 & 3  \\
\embedder{FAB-Ramy-v1} & 6 & 28 \\
\embedder{gte-base-en-v1.5} & 7 & 2  \\
\embedder{all-mpnet-base-v2} & 8 & 15 \\
\embedder{instructor-base} & 9 & 12 \\
\embedder{paraphrase-MiniLM-L12-v2} & 10 & 19 \\
\embedder{nomic-embed-text-v1.5} & 11 & 7  \\
\embedder{jina-embeddings-v2-base-en} & 12 & 10 \\
\embedder{MedEmbed-small-v0.1} & 13 & 8  \\
\embedder{stella-base-en-v2} & 14 & 6  \\
\embedder{Wartortle} & 15 & 22 \\
\embedder{LaBSE} & 16 & 24 \\
\embedder{all-MiniLM-L12-v2} & 17 & 17 \\
\embedder{gtr-t5-large} & 18 & 13 \\
\embedder{cde-small-v1} & 19 & 1  \\
\embedder{gte-micro} & 20 & 27 \\
\embedder{msmarco-bert-co-condensor} & 21 & 20 \\
\embedder{contriever-base-msmarco} & 22 & 18 \\
\embedder{snowflake} & 23 & 11 \\
\embedder{Ivysaur} & 24 & 14 \\
\embedder{Venusaur} & 25 & 23 \\
\embedder{distiluse-base-v2} & 26 & 29 \\
\embedder{allenai-specter} & 27 & 26 \\
\embedder{SGPT-125M} & 28 & 21 \\
\embedder{komninos} & 29 & 25 \\
         \bottomrule
    \end{tabular}}
    \caption{Comparing our obtained main ranking (Table~\ref{tab:mainres}) against the relative ranking on MTEB.}
    \label{tab:compare_ranking_mteb}
\end{table}

\end{document}